\pdfoutput=1

\documentclass{article}
\usepackage{spconf,amsmath,graphicx,hyperref}
\usepackage{amssymb}

\usepackage{amsmath}
\usepackage{booktabs}       
\usepackage[table]{xcolor}  
\usepackage{multirow}
\usepackage{amsfonts}

\usepackage{cite}

\title{SimToken: A Simple Baseline for Referring Audio-Visual Segmentation}

\name{Dian Jin$^{1,\ast}$ 
\quad Yanghao Zhou$^{3,\ast}$ 
\quad Jinxing Zhou$^{2,\dagger}$ 
\quad Jiaqi Ma$^{2}$ 
\quad Ruohao Guo$^{4}$ 
\quad Dan Guo$^{1,\dagger}$ 
\thanks{$^{\ast}$ These authors contributed equally to this work.}
\thanks{$^{\dagger}$ Corresponding authors (\{zhoujxhfut, guodandd\}@gmail.com).}
}

\address{
  $^{1}$ HFUT, Hefei, China\quad
  $^{2}$ MBZUAI, Abu Dhabi, UAE\quad
  $^{3}$ NUS, Singapore\quad
  $^{4}$ PKU, Beijing, China
}

\begin{document}
%
\maketitle
\begin{abstract}
Referring Audio-Visual Segmentation (Ref-AVS) aims to segment specific objects in videos based on natural language expressions involving audio, vision, and text information. This task poses significant challenges in cross-modal reasoning and fine-grained object localization. 
In this paper, we propose a simple framework, \textit{SimToken}, that integrates a multimodal large language model (MLLM) with the Segment Anything Model (SAM).
The MLLM is guided to generate a special semantic token representing the referred object. This compact token, enriched with contextual information from all modalities, acts as a prompt to guide SAM to segment objects across video frames.
To further improve semantic learning, we introduce a novel target-consistent semantic alignment loss that aligns token embeddings from different expressions but referring to the same object. 
Experiments on the Ref-AVS benchmark demonstrate that our approach achieves superior performance compared to existing methods. Code is \href{https://github.com/DianJin-HFUT/SimToken}{here}.

\end{abstract}
\begin{keywords}
Referring Audio-Visual Segmentation, Mutimodal Large Language Models, SAM.
\end{keywords}
\section{Introduction}
\label{sec:intro}

\begin{figure}[t]
    \centering
    \includegraphics[width=0.99\linewidth]{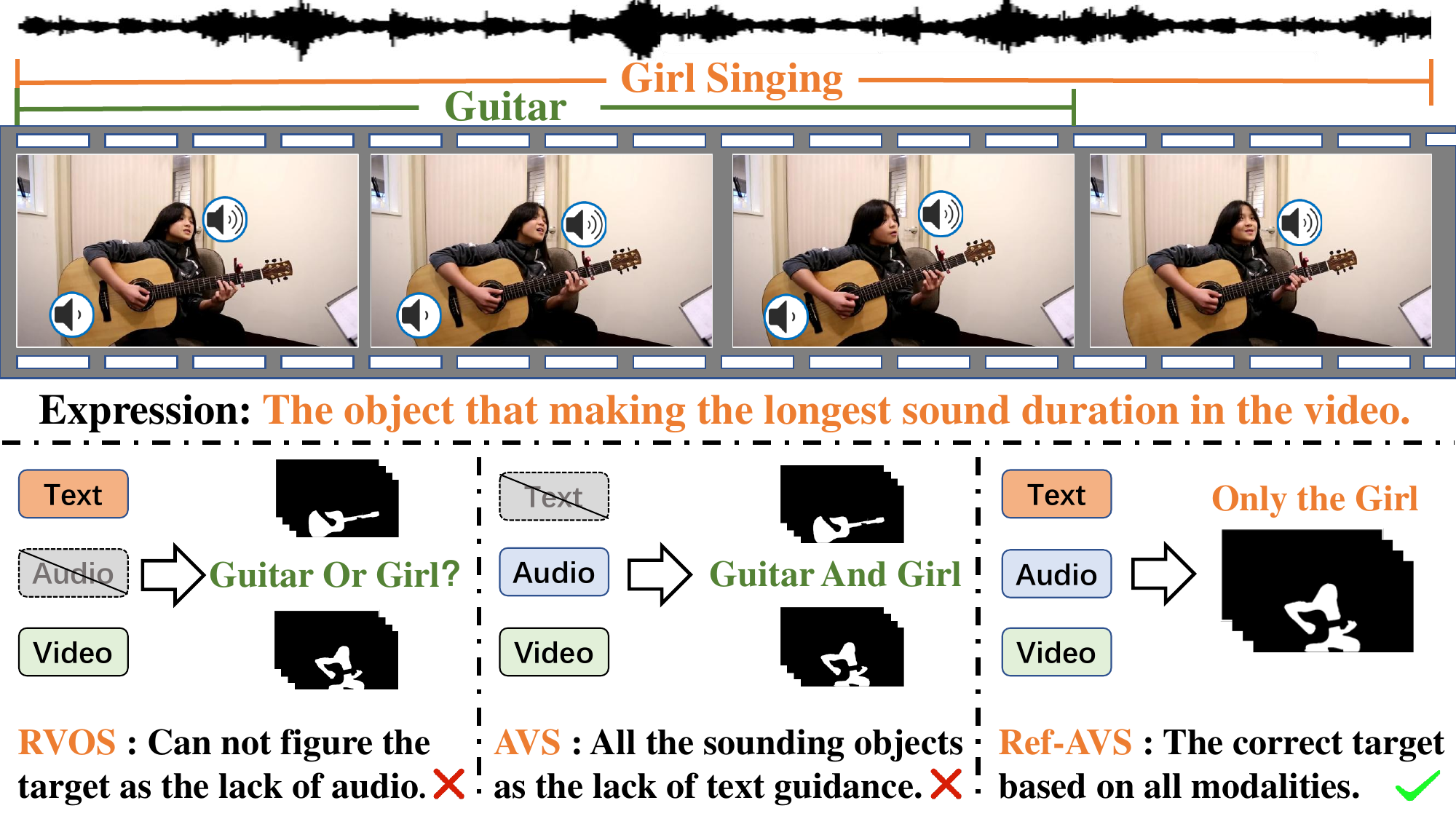}
    \vspace{-3ex}
    \caption{Comparison of Ref-AVS with AVS and RVOS tasks.}
    \vspace{-2ex}
    \label{fig:Introduction}
\end{figure}

Referring audio-visual segmentation (Ref-AVS)~\cite{wang2024ref} is an emerging task in audiovisual field~\cite{DBLP:journals/tomccap/ZhangLTWWZG25, 
DBLP:journals/tomccap/LiuXZLG25, 
DBLP:conf/aaai/LiZZTL025, 
DBLP:conf/aaai/ZhaoZZ0C25, 
DBLP:conf/aaai/ZhouZQTCG25, 
DBLP:conf/cvpr/ZhouGGMHZCW25, 
zhou2025clasp,  
DBLP:journals/ijcv/ZhouGZW24, 
DBLP:conf/aaai/LiGZZ024, 
DBLP:conf/eccv/ZhouGMZCW24, 
DBLP:conf/mm/MaoSZQZXZD24, 
DBLP:journals/pami/ZhouGW23, 
DBLP:conf/cvpr/ShenLZQHHLDKWQZ23, 
DBLP:conf/cvpr/ZhouZZH021} 
that aims to segment specific objects in videos based on audio, visual, and textual cues. As illustrated in Fig.~\ref{fig:Introduction}, unlike prior referring video object segmentation (RVOS)~\cite{ding2023mevis, 
wu2022language}, which uses only vision and language, or audio-visual segmentation (AVS)~\cite{DBLP:conf/eccv/ZhouWZSZBGKWZ22, DBLP:journals/ijcv/ZhouSWZSZBGKWZ25, DBLP:conf/cvpr/GuoYCNLQQZXY00Z25, zhou2025aloha, DBLP:journals/ijon/GuoHZ24, zhou2025mettle, zhou2025think}, which segments all sounding objects without text guidance, Ref-AVS enables selective segmentation based on both what is seen and heard, guided by natural language expressions.

Recent works~\cite{wang2024ref, radman2025tsam, wang2025sam2} have explored various strategies to integrate multimodal inputs for Ref-AVS. 
EEMC~\cite{wang2024ref} employs transformer-based fusion to combine audio, visual, and textual features. TSAM~\cite{radman2025tsam} and SAM2-LOVE~\cite{wang2025sam2} enhance segmentation by injecting multimodal prompts into foundation models such as SAM~\cite{kirillov2023segment} and SAM2~\cite{ravi2024sam}, thereby improving the quality of segmentation.
While existing methods have achieved promising results, a key challenge in the Ref-AVS task remains: how to more effectively understand and integrate information from multiple modalities, \textit{i.e.}, the audio, video, and textual expressions, for accurate segmentation?

\begin{figure*}[t]
    \centering
    \includegraphics[width=0.99\linewidth]{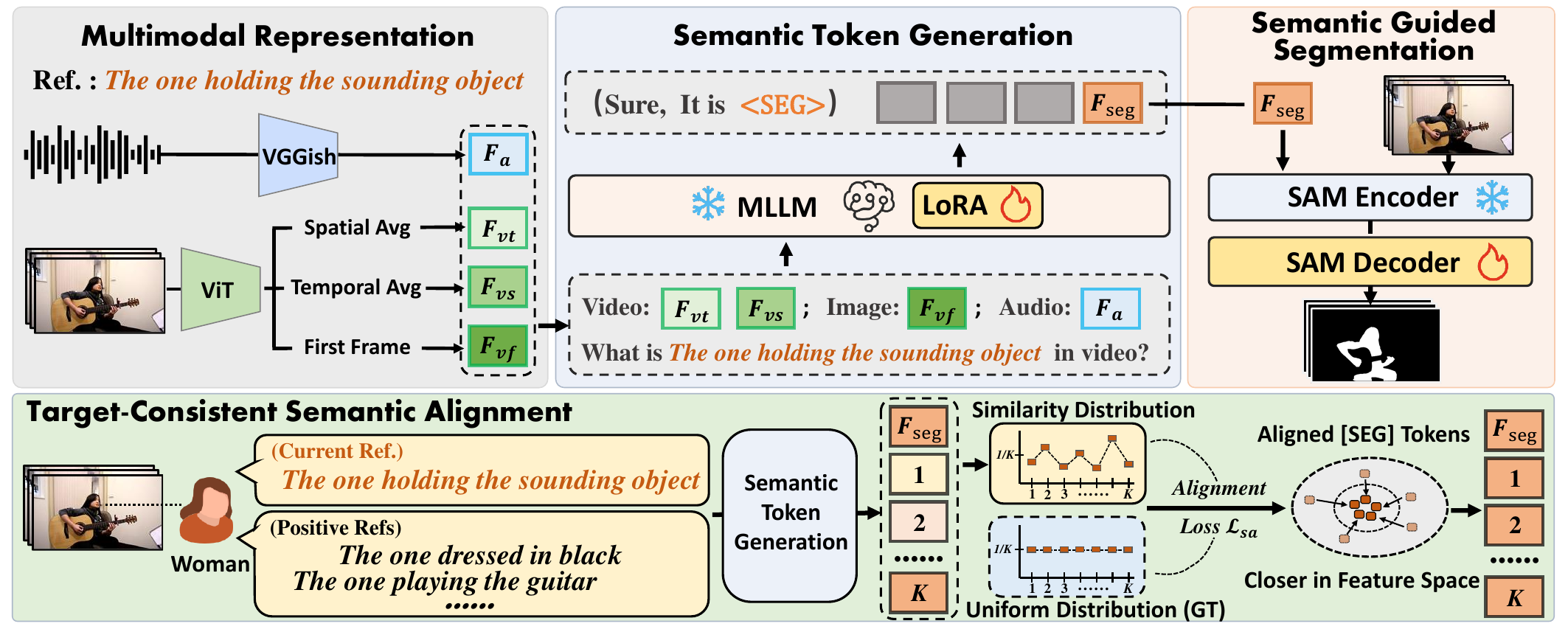}
    \vspace{-3ex}
    \caption{Framework Overview. 
    The multimodal signals are encoded
    and fed into a MLLM to generate a special semantic embedding \texttt{<SEG>} of the referred target object, which then prompt SAM model to perform segmentation across the video. An target-consistent semantic alignment loss is proposed to facilitate \texttt{<SEG>} token learning across varied referring expressions.
    }
    \vspace{-2ex}
    \label{fig:model}
\end{figure*}

Inspired by the strong capabilities of  Multimodal Large Language Model (MLLM) in cross-modal understanding and reasoning, we introduce MLLM into the Ref-AVS task.
The transformer-based MLLM usually suffers from huge computational cost due to the quadratic complexity.
In the Ref-AVS task, there are far more visual tokens than audio and textual tokens.
Therefore, we first design a multi-view visual token compression strategy, which adopts the first-frame preservation, spatial pooling, and temporal pooling.
Then, the multimodal visual features along the audio and reference text are inserted into a pre-defined instruction template and fed to the MLLM for cross-modal reasoning and response generation.
A special semantic token \texttt{<SEG>} is generated, which is expected to represent the referred object incorporating multimodal cues.
In this way, \texttt{<SEG>} can serve as an effective and unified signal to prompt the SAM foundation model to generate pixel-wise segmentation maps for all the video frames.
In addition, we observe that semantically diverse referring expressions can point to the same target. 
Motivated by this, we propose a \textit{Target-consistent Semantic Alignment} strategy, which brings the generated semantics of different expressions referring to the same object closer in the embedding space. This alignment further enhances the model's understanding and refines the semantic quality of generated \texttt{<SEG>} tokens.

In summary, our main contributions are three-fold: 
(1) We introduce the MLLM into the Ref-AVS task and integrate it with SAM, enabling high-quality, instruction-guided segmentation. We also improve the technique in visual token compression.
(2) We propose an alignment loss to enhance semantic consistency learning across diverse referring expressions that target the same object. 
(3) We conduct extensive experiments and demonstrate that our method achieves state-of-the-art performance on the Ref-AVSBench dataset.

\section{Methodology}
\label{sec:method}

\subsection{Semantic Token Generation}

The core challenge of the Ref-AVS task lies in understanding the relationships across modalities to accurately identify the target referred to by the expression.
The strong multimodal reasoning capabilities exhibited by Multimodal Large Language Models (MLLMs) make them suitable to address this issue, which have been explored in image~\cite{lai2024lisa} and video object segmentation~\cite{yan2024visa,du2025crab} tasks.
To reduce the computational cost introduced by large number of visual tokens, the most relevant work, Crab~\cite{du2025crab}, utilizes learnable Q-Former~\cite{li2023blip2} module to aggregate multiple video frames into several visual tokens.
In this work, we consider a parameter-free manner, which explicitly preserves the spatial and temporal visual information during the visual token compression. 

Specifically, given a $T$-second audible video, we first extract its visual feature $\mathbf{F}_v \in \mathbb{R}^{T \times L\times D}$ of video frames using a ViT-based encoder~\cite{dosovitskiy2020image}. Here, $L$ is the number of visual patches, and $D$ is the channel dimension.
Then we compress it into: $\mathbf{F}_{vt} \in \mathbb{R}^{T \times D}$ for temporal dynamics via spatial average pooling, $\mathbf{F}_{vs} \in \mathbb{R}^{L \times D}$ for global spatial layout via temporal average pooling.
To better preserve intrinsic spatial information, we keep the complete visual features of the first frame, denoted as $\mathbf{F}_{vf} \in \mathbb{R}^{L \times D}$. 
The audio feature $\mathbf{F}_{a} \in \mathbb{R}^{T \times D}$ is extracted by a pretrained VGGish~\cite{hershey2017cnn} model and the referring expression text is processed by a tokenizer.
Then, these multimodal features
are embedded into an instruction-style prompt and fed into the MLLM:
``\texttt{Video:$\mathbf{F}_{vt}, \mathbf{F}_{vs}$. Image:$\mathbf{F}_{vf}$. Audio:$\mathbf{F}_{a}$. What is \{referent\} in video?}''. 
We constrain the MLLM to always respond with: \texttt{It is <SEG>}. 
Through later fine-tuning, the model maps multimodal spatial-temporal cues into a compact semantic token \texttt{<SEG>}, whose embedding, denoted as $\mathbf{F}_{\text{seg}}$, captures category semantics of the referred target object and can be used across frames for subsequent segmentation.

\subsection{Semantic Guided Segmentation}

Once the target semantic token \texttt{<SEG>} is obtained, the next step is to convert it into high-quality segmentation masks. 
Instead of prompting the MLLM for each frame or using tracking-based mask propagation which increases cost and can accumulate errors, we leverage the fact that the MLLM receives global video information from inputs, including spatial-temporal cues and semantic features via $\mathbf{F}_{vt}$, $\mathbf{F}_{vs}$, and $\mathbf{F}_{vf}$. 
Thus, it is reasonable to assume that the MLLM can produce a unified target semantic representation that remains consistent across frames.
Based on this insight, we adopt a simple yet effective strategy: we generate the target semantics only once and use them to guide SAM for all video frames. This allows the model to learn frame-invariant semantics during training and enables efficient segmentation for the entire video sequence. The overall process can be summarized as:
\begin{equation}
\{m_t\}_{t=1}^T = \text{SAM} \left( \varepsilon_v(v_t), \varepsilon_p(\mathbf{F}_{\text{seg}}) \right),
\end{equation}
where $\varepsilon_v(\cdot)$ and $\varepsilon_p(\cdot)$ denote the visual and prompt encoders in SAM respectively, and $v_{t}$ denotes the $t$-th frame.
This strategy reduces inference cost, avoids temporal error propagation, and maintains spatial-temporal consistency without the need for explicit tracking.

\subsection{Target-Consistent Semantic Alignment}
As mentioned above, the MLLM is guided to generate special semantic token \texttt{<SEG>}, which then serves as guidance for the segmentation process.
\texttt{<SEG>} contains the category semantics of the target referred object.
In the studied Ref-AVS task, we observed that a single target object usually can be described from different perspectives.
As shown in Fig.~\ref{fig:model}, the target \textit{woman} can be referred to as ``\textit{the one holding the sounding object}'', ``\textit{the one dressed in black}'' and ``\textit{the one playing the guitar}''.
These varied referring expressions may be modality-specific (either audio or visual) or cross-modal, yet they all refer to the same target.
To ensure that the model produces consistent \texttt{<SEG>} outputs for the same target, we propose a novel semantic alignment loss.

Specifically, 
for the $i$-th video sample, 
let $\mathbf{q}_i$ be the semantic token embedding of \texttt{<SEG>} corresponding to a specific referring expression.
$\mathcal{P}_i =\{\mathbf{p}_1, \mathbf{p}_2, ..., \mathbf{p}_K\}$ is the positive semantic token embedding set where each element is produced from expression that refer to the same target as $\mathbf{q}_i$. $K$ is the total number of the positive instances for the $i$-th video sample.
We compute the similarity distribution of $\mathbf{q}_i$ over its positive set $\mathcal{P}_i$. Ideally, each semantic token in the positive set should be assigned equal importance, \textit{i.e.}, a uniform distribution
where every element has probability $\tfrac{1}{|\mathcal{P}_i|}$.
Therefore, the proposed semantic alignment loss $\mathcal{L}_{\text{sa}}$ is to regularize the cross-entropy between these two distributions, formulated as:
\begin{equation}
\mathcal{L}_{\text{sa}} = - \sum_{\mathbf{p} \in P_i}
\frac{1}
{|\mathcal{P}_i|} \log \frac{\exp\left( {\mathbf{q}_i^\top \mathbf{p}}/{\tau} \right)}
{\sum_{\mathbf{p'} \in \mathcal{P}_i} \exp\left( {\mathbf{q}_i^\top \mathbf{p'}}/{\tau} \right)} .
\end{equation}
Here, $\tau$=0.07 is the temperature hyperparameter.
By minimizing the loss, 
embeddings of different expressions referring to the same target are encouraged to converge toward a common semantic center, ensuring target-consistent representations that remain stable regardless of variations in the referring expressions.
Such alignment encourages the model to move beyond independently processing each input and instead to jointly consider the relationships among different expressions for the same target, thereby enhancing its overall understanding of the audio-visual scene. 
The final objective function is composed of $\mathcal{L}_\text{sa}$, the text auto-regression loss $\mathcal{L}_\text{text}$, and the segmentation loss $\mathcal{L}_\text{mask}$ between the predicted masks and the ground truth: $\mathcal{L}=\mathcal{L}_\text{text}+\mathcal{L}_\text{mask}+\lambda \mathcal{L}_\text{sa}$.

\section{Experiments}

\begin{table*}[t]
\centering
\begin{small}
\setlength{\tabcolsep}{8pt}
\begin{tabular}{l|c|ccc|ccc|ccc|c}
\toprule
\textbf{Method} & 
\textbf{Venue} &
\multicolumn{3}{c|}{\textbf{Seen $\uparrow$} } &
\multicolumn{3}{c|}{\textbf{Unseen $\uparrow$}} &
\multicolumn{3}{c|}{\textbf{Mix (S+U) $\uparrow$}} &
\textbf{Null} $\downarrow$ \\
& &
$\mathcal{J}$ & $\mathcal{F}$ & $\mathcal{J\&F}$ &
$\mathcal{J}$ & $\mathcal{F}$ & $\mathcal{J\&F}$ &
$\mathcal{J}$ & $\mathcal{F}$ & $\mathcal{J\&F}$ &
$\mathcal{S}$ \\
\midrule
EEMC~\cite{wang2024ref} & ECCV'2024 & 34.2 & 51.3 & 42.8 & 49.5 & 64.8 & 57.2 & 41.9 & 58.1 & 50.0 & \textbf{0.007}  \\
Crab~\cite{du2025crab} & CVPR'2025 & 40.5 & 58.0 & 49.3 & 45.6 & 63.0 & 54.3 & 43.1 & 60.5 & 46.2 & - \\
TSAM~\cite{radman2025tsam} & CVPR'2025 & 43.4 & 56.8 & 50.1 & 54.6 & 66.4 & 60.5 & 49.0 & 61.6 & 55.3 & 0.017  \\
SAM2-LOVE~\cite{wang2025sam2} \quad & CVPR'2025 & 43.5 & 51.9 & 47.7 & 66.5 & 72.3 & 69.4 & 55.0 & 62.1 & 58.5 & 0.23  \\

\midrule
\rowcolor{gray!10}
\textbf{SimToken (Ours)} & - & \textbf{72.0} & \textbf{81.3} & \textbf{76.7} & \textbf{69.8} & \textbf{79.1} & \textbf{74.5} & \textbf{70.9} & \textbf{80.2} & \textbf{75.6} & \underline{0.012} \\

\bottomrule
\end{tabular}
\end{small}
\vspace{-2ex}
\caption{Comparison with prior methods on Ref-AVSBench dataset. $\mathcal{J\&F} = (\mathcal{J}+\mathcal{F})/2$ in each subset.}
\vspace{-2ex}
\label{tab:main_comparison}
\end{table*}

\subsection{Experimental Setups}

We evaluate our method on the Ref-AVSBench dataset~\cite{wang2024ref}, which contains three subsets: 
Seen (categories seen during training), Unseen (novel categories), and Null (referring expressions with no valid target).
Following prior works~\cite{wang2024ref,wang2025sam2,radman2025tsam}, we use $\mathcal{J}$ and $\mathcal{F}$ for Seen and Unseen splits, and $S$ for Null, where higher $S$ indicates less accurate pixel grounding.

We adopt Chat-UniVi-7B~\cite{jin2024chat} as the pretrained MLLM, fine-tuning it via LoRA
 modules.
The audio embeddings are aligned with the MLLM embedding space using lightweight linear projector.
Notably, we do not use any domain-specific data for the projector pretraining, but jointly train it and the SAM decoder in an end-to-end manner using the proposed loss objective.
Training is conducted for 10 epochs using AdamW with a learning rate of $5 \times 10^{-5}$ and cosine decay.
The segmentation loss $\mathcal{L}_\text{mask}$ includes binary cross-entropy and Dice loss equally weighted. The loss balancing weight $\lambda$ is set to 0.1.
All experiments are conducted on an NVIDIA A800 80GB GPU.

\subsection{Comparison with State-of-the-Arts}
Table~\ref{tab:main_comparison} presents the quantitative comparison results on the Ref-AVSBench.
Our model achieves new state-of-the-art performance on the Seen, Unseen, and Mix test sets, significantly outperforming all previous Ref-AVS methods.
For example, our method surpasses SAM2-LOVE~\cite{wang2025sam2} by \textbf{29.0\%} and \textbf{5.1\%} in terms of $\mathcal{J\&F}$ metric on the Seen and Unseen sets, respectively.
Moreover, prior approaches using SAM or SAM2 usually exhibit low performance on the Null set with high $\mathcal{S}$.
Our method can achieve comparative performance as the previous best EEMC~\cite{wang2024ref}.
These results demonstrate the effectiveness and superiority of our proposed method.

\subsection{Ablation Study}

\begin{table}[t]
\centering
\resizebox{\linewidth}{!}{%
\begin{tabular}{lccccccc}
\toprule
\multirow{2}{*}{Setting} & \multicolumn{2}{c}{Seen} & \multicolumn{2}{c}{Unseen} & \multicolumn{2}{c}{Mix} & \multicolumn{1}{c}{Null} \\
\cmidrule(lr){2-3} \cmidrule(lr){4-5} \cmidrule(lr){6-7} \cmidrule(lr){8-8}
& $\mathcal{J}$ & $\mathcal{F}$ & $\mathcal{J}$ & $\mathcal{F}$ & $\mathcal{J}$ & $\mathcal{F}$ & $\mathcal{S}$ \\
\midrule
Full & \textbf{72.0} & \textbf{81.3} & \textbf{69.8} & \textbf{79.1} & \textbf{70.9} & \textbf{80.2} & \textbf{0.012} \\
w/o A & 69.2 & 79.4 & 68.0 & 78.6 & 68.6 & 79.0 & 0.102 \\
w/o A\&V \quad \quad & 59.9 & 72.6 & 63.8 & 75.1 & 61.9 & 73.8 & 0.079 \\
\bottomrule
\end{tabular}
}
\vspace{-3ex}
\caption{Ablation study on the input modalities to the MLLM. }
\label{tab:ablation_modalities}
\end{table}

\noindent\textbf{Ablation Study on Modalities.}
The Ref-AVS task requires models to understand multiple modalities in order to accurately segment the referred target. 
We conduct an ablation study by selectively removing individual modalities from the input to explore their impacts.
As shown in Table~\ref{tab:ablation_modalities}, removing the audio input leads to a performance drop on both the Seen and Unseen sets. More importantly, the $\mathcal{S}$ score on the Null set increases significantly, indicating a higher rate of false positives in scenarios where no target is present. This suggests that audio information is crucial not only for accurate segmentation but also for correctly identifying cases where segmentation should not occur. When we further remove the visual input, the segmentation performance degrades dramatically across all test sets.
These results confirm that our approach effectively utilizes all modalities and does not rely on any single modality for biased or partial reasoning.

\begin{table}[t]
\centering
\resizebox{\linewidth}{!}{%
\begin{tabular}{lccccccc}
\toprule
\multirow{2}{*}{Setting} & \multicolumn{2}{c}{Seen} & \multicolumn{2}{c}{Unseen} & \multicolumn{2}{c}{Mix} & \multicolumn{1}{c}{Null} \\
\cmidrule(lr){2-3} \cmidrule(lr){4-5} \cmidrule(lr){6-7} \cmidrule(lr){8-8}
& $\mathcal{J}$ & $\mathcal{F}$ & $\mathcal{J}$ & $\mathcal{F}$ & $\mathcal{J}$ & $\mathcal{F}$ & $\mathcal{S}$ \\
\midrule
$\mathbf{F}_{vf}$ & 70.7 & 80.0 & 65.4 & 75.8 & 68.0 & 77.9 & 0.012 \\
$\mathbf{F}_{vf},\mathbf{F}_{vs}$ & 71.1 & 80.7 & 69.1 & 78.3 & 70.1 & 80.0 & 0.012 \\
$\mathbf{F}_{vf},\mathbf{F}_{vs},\mathbf{F}_{vt}$ & \textbf{72.0} & \textbf{81.3} & \textbf{69.8} & \textbf{79.1} & \textbf{70.9} & \textbf{80.2} & \textbf{0.012} \\
\bottomrule
\end{tabular}
}
\vspace{-3ex}
\caption{Ablation study on the visual feature expressions.}
\label{tab:ablation_videofeature}
\end{table}

\noindent\textbf{Impact of Visual Feature Expression.}
In our method, we provide the model with spatial-temporal context through 
\( \mathbf{F}_{vt} \), \( \mathbf{F}_{vs} \), and \( \mathbf{F}_{vf} \), rather than feeding all video frames. 
To assess the effectiveness of these components in helping the model understand the video content, we conduct an ablation study across three settings: 
(1) using only the single-frame feature \( \mathbf{F}_{vf} \); 
(2) further incorporating  \( \mathbf{F}_{vs} \);
(3) using the full visual features $\mathbf{F}_{vf}+\mathbf{F}_{vs}+\mathbf{F}_{vt}$.
As shown in Table~\ref{tab:ablation_videofeature}, 
using only $\mathbf{F}_{vf}$, the model already achieves high performance especially for the Seen test set.
Introducing \( \mathbf{F}_{vs} \) significantly improves the performance for Unseen set.
Combining all three types of visual features, the model gains further improvements and reaches the highest performance.
These results demonstrate the benefits of each visual component, where $\mathbf{F}_{vf}$ provides a raw image reference, and $\mathbf{F}_{vs}$ and $\mathbf{F}_{vt}$ encodes the spatial and temporal dynamics across frames.

\begin{table}[t]
\centering
\resizebox{\linewidth}{!}{%
\begin{tabular}{lccccccc}
\toprule
\multirow{2}{*}{Setting} & \multicolumn{2}{c}{Seen} & \multicolumn{2}{c}{Unseen} & \multicolumn{2}{c}{Mix} & \multicolumn{1}{c}{Null} \\
\cmidrule(lr){2-3} \cmidrule(lr){4-5} \cmidrule(lr){6-7} \cmidrule(lr){8-8}
& $\mathcal{J}$ & $\mathcal{F}$ & $\mathcal{J}$ & $\mathcal{F}$ & $\mathcal{J}$ & $\mathcal{F}$ & $\mathcal{S}$ \\
\midrule
w/o $\mathcal{L}_{\text{sa}}$ \quad \quad \quad& 70.8 & 80.7 & 66.9 & 76.7 & 68.8 & 78.7 & {0.012} \\
w/ $\mathcal{L}_{\text{sa}}$ & \textbf{72.0} & \textbf{81.3} & \textbf{69.8} & \textbf{79.1} & \textbf{70.9} & \textbf{80.2} & \textbf{0.012} \\
\bottomrule
\end{tabular}
}
\vspace{-3ex}
\caption{Ablation study on the semantic alignment loss $\mathcal{L}_{\text{sa}}$ .}
\label{tab:ablation_align}
\end{table}

\noindent\textbf{Effectiveness of the Alignment Loss $\mathcal{L}_{\text{sa}}$.}
We conduct an ablation study by training the model with or without this loss item.
As shown in Table~\ref{tab:ablation_align}, introducing the alignment loss leads to consistent and notable performance improvements across all test sets. 
The result indicates that such alignment enables the MLLM to generate more semantically consistent outputs across different expressions for the same target. 
By establishing cross-reference associations, the model is further enhanced to understand audio-visual scenes.

\subsection{Qualitative Results}
\begin{figure}[t]
    \centering
    \includegraphics[width=1.0\linewidth]{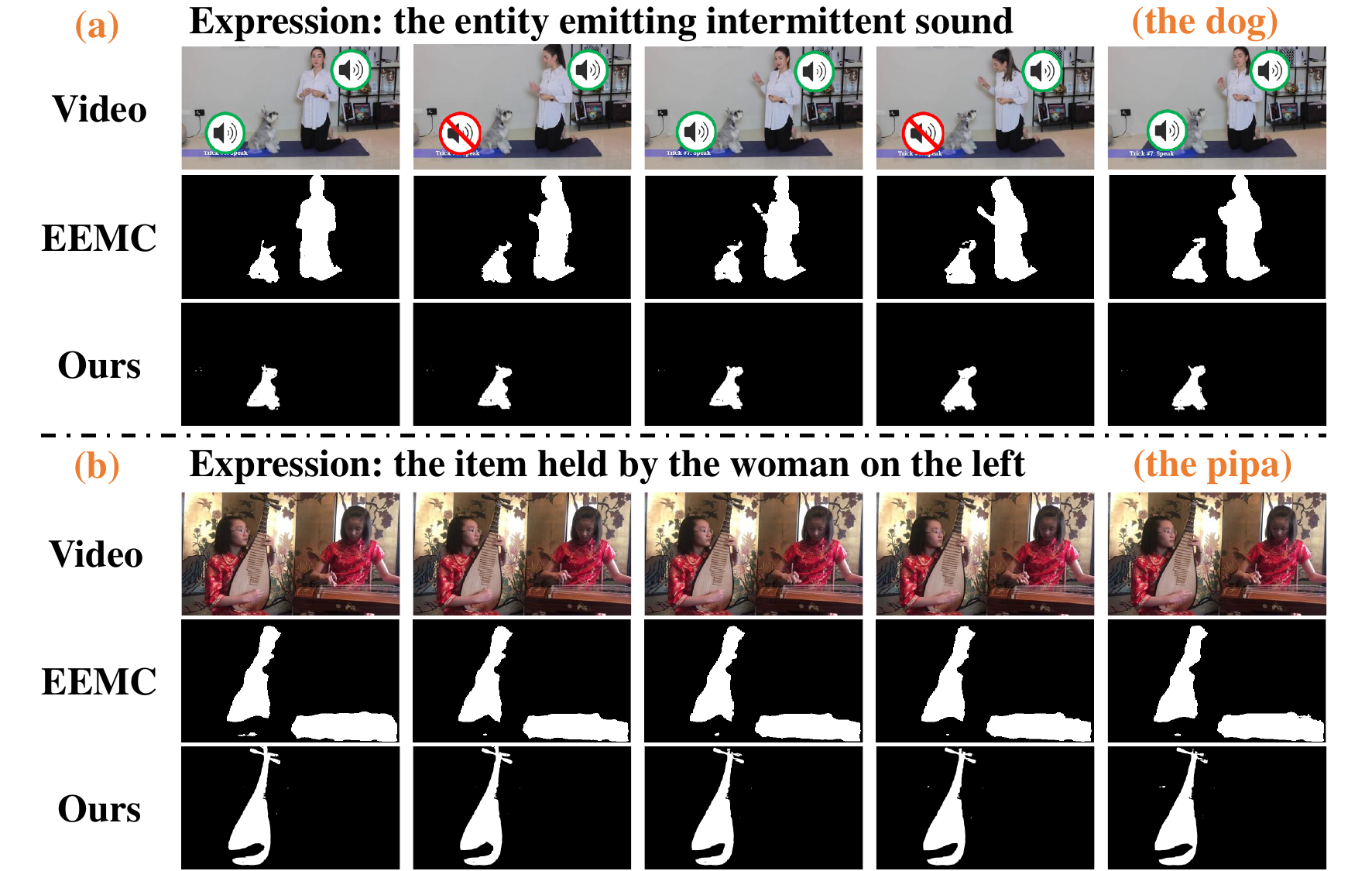}
    \vspace{-5.5ex}
    \caption{Visualization of the predicted segmentation maps.}
    \vspace{-1ex}
    \label{fig:visualization}
\end{figure}

To provide an intuitive understanding of our model’s performance, we visualize predicted segmentation results and compare them with those produced by the baseline method EEMC~\cite{wang2024ref}. As shown in Fig.~\ref{fig:visualization}, our model performs well in challenging cases that require global temporal reasoning (a) and spatial reasoning (b).
EEMC tends to produce inaccurate segmentation maps which contains irrelevant objects, \textit{e.g.}, the \textit{woman} and \textit{guzheng}.
In contrast, our method accurately identifies the referred target objects and produces high-quality masks.
These examples highlight the superiority of our method, which correctly understands the referring expressions by effectively analyzing spatial-temporal cues from multiple modalities.

\section{Conclusion}
We present SimToken, a simple yet effective semantic-token based baseline for the Ref-AVS task.
Our method utilizes a pretrained MLLM to process multimodal signals, perform spatial-temporal reasoning, and generate compact embeddings of the referred objects, which guides the SAM model to generate accurate segmentation maps.
We also introduce a target-consistent alignment loss to enhance semantic learning by exploring relations among multiple referring expressions.
Our framework is simple and unified, requires no complex pretraining, yet achieves competitive performance on Ref-AVSBench dataset, establishing a strong MLLM-based baseline for future research.


\ninept
\bibliographystyle{IEEEbib}


\begin{thebibliography}{10}

\bibitem{wang2024ref}
Y~Wang, P~Sun, D~Zhou, G~Li, H~Zhang, and D~Hu,
\newblock ``Ref-avs: Refer and segment objects in audio-visual scenes,''
\newblock in {\em ECCV}, 2024, pp. 196--213.

\bibitem{DBLP:journals/tomccap/ZhangLTWWZG25}
Z~Zhang, K~Li, S~Tang, Y~Wei, F~Wang, J~Zhou, and D~Guo,
\newblock ``Temporal boundary awareness network for repetitive action counting,''
\newblock {\em {ACM} TOMM}, pp. 118:1--118:22, 2025.

\bibitem{DBLP:journals/tomccap/LiuXZLG25}
X~Liu, N~Xia, J~Zhou, Z~Li, and D~Guo,
\newblock ``Towards energy-efficient audio-visual classification via multimodal interactive spiking neural network,''
\newblock {\em {ACM} TOMM}, pp. 1--24, 2025.

\bibitem{DBLP:conf/aaai/LiZZTL025}
Z~Li, J~Zhou, J~Zhang, S~Tang, K~Li, and D~Guo,
\newblock ``Patch-level sounding object tracking for audio-visual question answering,''
\newblock in {\em {AAAI}}, 2025, pp. 5075--5083.

\bibitem{DBLP:conf/aaai/ZhaoZZ0C25}
P~Zhao, J~Zhou, Y~Zhao, D~Guo, and Y~Chen,
\newblock ``Multimodal class-aware semantic enhancement network for audio-visual video parsing,''
\newblock in {\em {AAAI}}, 2025, pp. 10448--10456.

\bibitem{DBLP:conf/aaai/ZhouZQTCG25}
Z~Zhou, J~Zhou, W~Qian, S~Tang, X~Chang, and D~Guo,
\newblock ``Dense audio-visual event localization under cross-modal consistency and multi-temporal granularity collaboration,''
\newblock in {\em {AAAI}}, 2025, pp. 10905--10913.

\bibitem{DBLP:conf/cvpr/ZhouGGMHZCW25}
J~Zhou, D~Guo, R~Guo, Y~Mao, J~Hu, Y~Zhong, X~Chang, and M~Wang,
\newblock ``Towards open-vocabulary audio-visual event localization,''
\newblock in {\em {CVPR}}, 2025, pp. 8362--8371.

\bibitem{zhou2025clasp}
J~Zhou, Z~Zhou, Y~Zhou, Y~Mao, Z~Duan, and D~Guo,
\newblock ``Clasp: Cross-modal salient anchor-based semantic propagation for weakly-supervised dense audio-visual event localization,''
\newblock {\em arXiv preprint arXiv:2508.04566}, 2025.

\bibitem{DBLP:journals/ijcv/ZhouGZW24}
J~Zhou, D~Guo, Y~Zhong, and M~Wang,
\newblock ``Advancing weakly-supervised audio-visual video parsing via segment-wise pseudo labeling,''
\newblock {\em IJCV}, pp. 5308--5329, 2024.

\bibitem{DBLP:conf/aaai/LiGZZ024}
Z~Li, D~Guo, J~Zhou, J~Zhang, and M~Wang,
\newblock ``Object-aware adaptive-positivity learning for audio-visual question answering,''
\newblock in {\em {AAAI}}, 2024, pp. 3306--3314.

\bibitem{DBLP:conf/eccv/ZhouGMZCW24}
J~Zhou, D~Guo, Y~Mao, Y~Zhong, X~Chang, and Meng Wang,
\newblock ``Label-anticipated event disentanglement for audio-visual video parsing,''
\newblock in {\em {ECCV}}, 2024, pp. 35--51.

\bibitem{DBLP:conf/mm/MaoSZQZXZD24}
Y~Mao, X~Shen, J~Zhang, Z~Qin, J~Zhou, M~Xiang, Y~Zhong, and Y~Dai,
\newblock ``Tavgbench: Benchmarking text to audible-video generation,''
\newblock in {\em {ACM} Multimedia}, 2024, pp. 6607--6616.

\bibitem{DBLP:journals/pami/ZhouGW23}
J~Zhou, D~Guo, and M~Wang,
\newblock ``Contrastive positive sample propagation along the audio-visual event line,''
\newblock {\em {IEEE} Trans. Pattern Anal. Mach. Intell.}, pp. 7239--7257, 2023.

\bibitem{DBLP:conf/cvpr/ShenLZQHHLDKWQZ23}
X~Shen, D~Li, J~Zhou, Z~Qin, B~He, X~Han, A~Li, Y~Dai, L~Kong, M~Wang, Y~Qiao, and Y~Zhong,
\newblock ``Fine-grained audible video description,''
\newblock in {\em {CVPR}}, 2023, pp. 10585--10596.

\bibitem{DBLP:conf/cvpr/ZhouZZH021}
J~Zhou, L~Zheng, Y~Zhong, S~Hao, and M~Wang,
\newblock ``Positive sample propagation along the audio-visual event line,''
\newblock in {\em {CVPR}}, 2021, pp. 8436--8444.

\bibitem{ding2023mevis}
H~Ding, C~Liu, S~He, X~Jiang, and C~C Loy,
\newblock ``Mevis: A large-scale benchmark for video segmentation with motion expressions,''
\newblock in {\em ICCV}, 2023, pp. 2694--2703.

\bibitem{wu2022language}
J~Wu, Y~Jiang, P~Sun, Z~Yuan, and P~Luo,
\newblock ``Language as queries for referring video object segmentation,''
\newblock in {\em CVPR}, 2022, pp. 4974--4984.

\bibitem{DBLP:conf/eccv/ZhouWZSZBGKWZ22}
J~Zhou, J~Wang, J~Zhang, W~Sun, J~Zhang, Stan Birchfield, D~Guo, L~Kong, M~Wang, and Y~Zhong,
\newblock ``Audio-visual segmentation,''
\newblock in {\em {ECCV} {(37)}}, 2022, pp. 386--403.

\bibitem{DBLP:journals/ijcv/ZhouSWZSZBGKWZ25}
J~Zhou, X~Shen, J~Wang, J~Zhang, W~Sun, J~Zhang, S~Birchfield, D~Guo, L~Kong, M~Wang, and Y~Zhong,
\newblock ``Audio-visual segmentation with semantics,''
\newblock {\em IJCV}, pp. 1644--1664, 2025.

\bibitem{DBLP:conf/cvpr/GuoYCNLQQZXY00Z25}
R~Guo, X~Ying, Y~Chen, D~Niu, G~Li, L~Qu, Y~Qi, J~Zhou, B~Xing, W~Yue, J~Shi, Q~Wang, P~Zhang, and B~Liang,
\newblock ``Audio-visual instance segmentation,''
\newblock in {\em {CVPR}}, 2025, pp. 13550--13560.

\bibitem{zhou2025aloha}
Y~Zhou, H~Huang, C~Guo, R~Tu, Z~Xiao, B~Wang, and X~Mao,
\newblock ``Aloha: Adapting local spatio-temporal context to enhance the audio-visual semantic segmentation,''
\newblock {\em {ACM} TOMM}, pp. 1--23, 2025.

\bibitem{DBLP:journals/ijon/GuoHZ24}
C~Guo, H~Huang, and Y~Zhou,
\newblock ``Enhance audio-visual segmentation with hierarchical encoder and audio guidance,''
\newblock {\em Neurocomputing}, p. 127885, 2024.

\bibitem{zhou2025mettle}
J~Zhou, Z~Li, Y~Yu, Y~Zhou, R~Guo, G~Li, Y~Mao, M~Han, X~Chang, and M~Wang,
\newblock ``Mettle: Meta-token learning for memory-efficient audio-visual adaptation,''
\newblock {\em arXiv preprint arXiv:2506.23271}, 2025.

\bibitem{zhou2025think}
J~Zhou, Y~Zhou, M~Han, T~Wang, X~Chang, H~Cholakkal, and R~M Anwer,
\newblock ``Think before you segment: An object-aware reasoning agent for referring audio-visual segmentation,''
\newblock {\em arXiv preprint arXiv:2508.04418}, 2025.

\bibitem{radman2025tsam}
A~Radman and J~Laaksonen,
\newblock ``Tsam: Temporal sam augmented with multimodal prompts for referring audio-visual segmentation,''
\newblock in {\em CVPR}, 2025, pp. 23947--23956.

\bibitem{wang2025sam2}
Y~Wang, H~Xu, Y~Liu, J~Li, and Y~Tang,
\newblock ``Sam2-love: Segment anything model 2 in language-aided audio-visual scenes,''
\newblock in {\em CVPR}, 2025, pp. 28932--28941.

\bibitem{kirillov2023segment}
A~Kirillov, E~Mintun, N~Ravi, H~Mao, C~Rolland, L~Gustafson, T~Xiao, S~Whitehead, A~C Berg, W~Lo, et~al.,
\newblock ``Segment anything,''
\newblock in {\em ICCV}, 2023, pp. 4015--4026.

\bibitem{ravi2024sam}
N~Ravi, V~Gabeur, Y~Hu, R~Hu, C~Ryali, T~Ma, H~Khedr, R~R{\"a}dle, C~Rolland, L~Gustafson, et~al.,
\newblock ``Sam 2: Segment anything in images and videos,''
\newblock {\em arXiv preprint arXiv:2408.00714}, 2024.

\bibitem{lai2024lisa}
X~Lai, Z~Tian, Y~Chen, Y~Li, Y~Yuan, S~Liu, and J~Jia,
\newblock ``Lisa: Reasoning segmentation via large language model,''
\newblock in {\em CVPR}, 2024, pp. 9579--9589.

\bibitem{yan2024visa}
C~Yan, H~Wang, S~Yan, X~Jiang, Y~Hu, G~Kang, W~Xie, and E~Gavves,
\newblock ``Visa: Reasoning video object segmentation via large language models,''
\newblock in {\em ECCV}, 2024, pp. 98--115.

\bibitem{du2025crab}
H~Du, G~Li, C~Zhou, C~Zhang, A~Zhao, and D~Hu,
\newblock ``Crab: A unified audio-visual scene understanding model with explicit cooperation,''
\newblock in {\em CVPR}, 2025, pp. 18804--18814.

\bibitem{li2023blip2}
J~Li, D~Li, S~Savarese, and S~Hoi,
\newblock ``Blip-2: Bootstrapping language-image pre-training with frozen image encoders and large language models,''
\newblock in {\em ICML}, 2023, pp. 19730--19742.

\bibitem{dosovitskiy2020image}
A~Dosovitskiy, L~Beyer, A~Kolesnikov, D~Weissenborn, X~Zhai, T~Unterthiner, M~Dehghani, M~Minderer, G~Heigold, S~Gelly, et~al.,
\newblock ``An image is worth 16x16 words: Transformers for image recognition at scale,''
\newblock {\em arXiv preprint arXiv:2010.11929}, 2020.

\bibitem{hershey2017cnn}
S~Hershey, S~Chaudhuri, D~PW Ellis, J~F Gemmeke, A~Jansen, R~C Moore, M~Plakal, D~Platt, R~Saurous, B~Seybold, et~al.,
\newblock ``Cnn architectures for large-scale audio classification,''
\newblock in {\em ICASSP}, 2017, pp. 131--135.

\bibitem{jin2024chat}
P~Jin, R~Takanobu, W~Zhang, X~Cao, and L~Yuan,
\newblock ``Chat-univi: Unified visual representation empowers large language models with image and video understanding,''
\newblock in {\em CVPR}, 2024, pp. 13700--13710.

\end{thebibliography}

\end{document}